\documentclass[runningheads]{llncs}
\usepackage{graphicx}
\usepackage{booktabs}
\usepackage{caption}
\usepackage{subcaption}
\usepackage{tikz}
\usepackage{pgfplots}
\pgfplotsset{compat=1.18}
\usepackage{algorithm}
\usepackage{algorithmic}
\usepackage{amsmath}

\begin{document}

\title{Decentralizing AI Memory: SHIMI, a Semantic Hierarchical Memory Index for Scalable Agent Reasoning\thanks{Supported by Concia}}

\author{Tooraj Helmi\orcidID{0009-0000-4959-1735}}

\institute{University of Southern California, Los Angeles, CA, USA \\
\email{thelmi@usc.edu}}

\maketitle              

\begin{abstract}
Retrieval-Augmented Generation (RAG) and vector-based search have become foundational tools for memory in AI systems, yet they struggle with abstraction, scalability, and semantic precision—especially in decentralized environments. We present SHIMI (Semantic Hierarchical Memory Index), a unified architecture that models knowledge as a dynamically structured hierarchy of concepts, enabling agents to retrieve information based on meaning rather than surface similarity. SHIMI organizes memory into layered semantic nodes and supports top-down traversal from abstract intent to specific entities, offering more precise and explainable retrieval. Critically, SHIMI is natively designed for decentralized ecosystems, where agents maintain local memory trees and synchronize them asynchronously across networks. We introduce a lightweight sync protocol that leverages Merkle-DAG summaries, Bloom filters, and CRDT-style conflict resolution to enable partial synchronization with minimal overhead. Through benchmark experiments and use cases involving decentralized agent collaboration, we demonstrate SHIMI’s advantages in retrieval accuracy, semantic fidelity, and scalability—positioning it as a core infrastructure layer for decentralized cognitive systems.


\end{abstract}
\section{Introduction}

As decentralized computing moves beyond financial transactions into cognitive infrastructure, there is growing demand for intelligent agents capable of reasoning, adaptation, and autonomous coordination. From decentralized autonomous organizations (DAOs)~\cite{hassan2021decentralized} to blockchain-native AI agents~\cite{liu2021blockchainagents}, these systems require not only execution logic, but also access to structured, interpretable memory. Agents must retrieve knowledge, match tasks, and share context across evolving environments—yet most memory systems in AI today are neither explainable~\cite{liebherr2025working} nor decentralized ~\cite{zheng2024decentralized}.

Existing retrieval methods, such as dense vector search and Retrieval-Augmented Generation (RAG)~\cite{lewis2020retrieval}, are limited in three ways. First, they rely on flat, unstructured embeddings that cannot express conceptual abstraction or compositional meaning. Second, their reliance on centralized indices conflicts with decentralized trust and infrastructure models. Third, their retrieval results are often difficult to interpret or justify—especially when the system operates autonomously or interacts with multiple peers.

To address these limitations, we introduce \textbf{SHIMI} (Semantic Hierarchical Memory Index), a new memory architecture designed for decentralized AI systems. Inspired by cognitive theories of abstraction~\cite{tulving1972organization}, SHIMI models memory as a dynamic tree of semantic concepts. Agents retrieve knowledge by descending from abstract goals toward specific entities, enabling meaning-based matching and interpretable reasoning paths.

Unlike RAG and similar approaches, SHIMI embeds semantics into both its retrieval and synchronization layers. Each agent maintains its own local semantic tree and updates it independently. Synchronization across agents is handled through a hybrid protocol using Merkle-DAG summaries~\cite{benet2014ipfs}, Bloom filters~\cite{bloom1970space}, and CRDT-style merging~\cite{shapiro2011conflict}. This allows partial, low-bandwidth updates while preserving convergence and consistency across asynchronous networks.

We evaluate SHIMI in a simulated task assignment scenario involving multiple agents. Our results show that SHIMI improves retrieval accuracy and interpretability while significantly reducing synchronization cost. These properties position SHIMI as a foundational component for decentralized memory in emerging AI-native infrastructures. As contributions, in this paper, we:

\begin{enumerate}
    \item Propose SHIMI, a semantic hierarchical memory structure for decentralized AI reasoning;
    \item Design a low-bandwidth synchronization protocol using Merkle-DAGs, Bloom filters, and CRDTs;
    \item Evaluate SHIMI against vector-based baselines across accuracy, traversal efficiency, and sync overhead;
    \item Demonstrate SHIMI’s generalizability across agent matching, decentralized knowledge indexing, and memory synchronization use cases.
\end{enumerate}

\section{Background and Related Work}

In this section, we review existing approaches to knowledge retrieval in AI systems, discuss their limitations in decentralized environments, and outline the foundational technologies that influence SHIMI's design.

\subsection{Retrieval-Augmented Generation and Vector Search}

Retrieval-Augmented Generation (RAG)~\cite{lewis2020retrieval} augments language models with external memory by retrieving passages relevant to a given query and feeding them as additional context into the model. Most RAG implementations use dense vector representations and rely on embedding similarity to identify relevant content, with pretrained models such as BERT~\cite{devlin2018bert} and Sentence-BERT~\cite{reimers2019sentence} commonly used to generate these embeddings. Libraries like FAISS~\cite{johnson2019billion} or Annoy~\cite{spotify_annoy} support scalable approximate nearest-neighbor search over millions of items.

While vector-based retrieval systems have demonstrated effectiveness in centralized applications, they exhibit several limitations that hinder their broader applicability. Firstly, these systems often employ a flat memory structure without inherent hierarchy or abstraction, complicating the modeling of conceptual relationships and impeding nuanced reasoning over data \cite{taipalus2023vector}. Secondly, embedding-based approaches are susceptible to semantic drift, where retrieved items may be lexically similar but semantically irrelevant, leading to decreased retrieval precision and potential inaccuracies in outputs \cite{gao2021complementing}. Additionally, the decision-making processes in vector-based retrieval often operate as "black boxes," offering limited transparency and making it challenging to provide human-aligned justifications for retrieval outcomes \cite{lakkaraju2017interpretable}. Furthermore, many implementations rely on a single index within a centralized infrastructure, posing challenges for scalability and limiting applicability in decentralized environments where data distribution is essential \cite{sharma2019certifai}. Addressing these limitations is crucial for enhancing the performance and applicability of vector-based retrieval systems, particularly in contexts requiring explainability, semantic precision, and decentralized operation

\subsection{Semantic, Hierarchical, and Graph-Based Memory Structures}

Memory systems inspired by human cognition typically organize knowledge across layers of abstraction~\cite{tulving1972organization}. These structures allow for flexible generalization, partial recall, and top-down conceptual refinement. Ontology-based approaches and semantic web frameworks~\cite{gruber1995toward} aim to formalize such organization using structured schemas, but often require extensive manual curation.

Graph-based retrieval systems such as DBpedia~\cite{auer2007dbpedia}, ConceptNet~\cite{speer2017conceptnet}, and Wikidata~\cite{vrandevcic2014wikidata} represent knowledge as nodes and labeled edges, allowing relational reasoning via graph traversal. More recently, graph neural networks (GNNs)~\cite{wu2020comprehensive} have been used to propagate contextual information across nodes, enabling deep semantic learning. However, these systems face challenges with scalability, schema rigidity, and centralization. Furthermore, they lack native support for decentralized synchronization or agent-local updates.

Related work has also explored hierarchical embedding spaces~\cite{vendrov2015order} and memory-augmented neural networks~\cite{graves2016hybrid}, but these models typically operate in closed environments with static or preloaded data.

\subsection{Memory in Decentralized Systems}

Decentralized storage frameworks such as IPFS~\cite{benet2014ipfs}, OrbitDB~\cite{orbitdb2021}, and peer-to-peer knowledge networks~\cite{fensel2021knowledge} provide alternatives to centralized data storage. However, these systems generally offer content-addressed persistence rather than meaning-aware memory. They are not designed to support retrieval by conceptual relevance or abstraction level.

To address the challenge of decentralized memory synchronization, several primitives have proven useful. Merkle-DAGs~\cite{benet2014ipfs} enable efficient structural hashing and diffing. Bloom filters~\cite{bloom1970space} provide compact probabilistic summaries of large sets. CRDTs (Conflict-Free Replicated Data Types)~\cite{shapiro2011conflict} allow distributed systems to converge on a shared state without global coordination. These components form the building blocks of SHIMI’s decentralized synchronization protocol.

\subsection{Positioning SHIMI}

To our knowledge, SHIMI is the first system to unify semantic abstraction, hierarchical memory organization, and decentralized synchronization into a single retrieval architecture. It addresses the shortcomings of flat vector search, static graphs, and centralized indexes by enabling agents to retrieve information meaningfully, efficiently, and independently—while ensuring eventual convergence across a distributed network.

\section{Approach}

SHIMI (\textit{Semantic Hierarchical Memory Index}) is a memory architecture designed to enable meaning-driven retrieval in decentralized agent systems. It combines a semantically structured memory model with a lightweight, scalable synchronization protocol. In this section, we describe both aspects in detail.

\subsection{Semantic Memory Architecture}

SHIMI organizes memory as a dynamic tree structure in which semantic abstractions are layered hierarchically. Formally, the memory is modeled as a rooted directed tree \( T = (V, E) \), where each node \( v \in V \) represents a semantic concept, and each directed edge \( (v_i, v_j) \in E \) encodes a parent-child relationship. Nodes may store entities, act as abstractions, or dynamically adapt their structure through merging.

We define the following parameters, aligned with our implementation: \( T \), the maximum number of child nodes under a parent (tree branching factor); \( L \), the maximum number of abstraction levels used for generalization; \( \delta \in [0,1] \), the similarity threshold for placement or matching; and \( \gamma \in [0,1] \), the compression ratio for semantic abstraction (word reduction).

Each node \( v \in V \) contains: a semantic summary \( s(v) \in \mathcal{L} \); a list of children \( C(v) \subseteq V \); a set of entities \( \mathcal{E}_v \subseteq \mathcal{E} \); and a parent pointer \( p(v) \in V \cup \{\bot\} \).

The core operation in SHIMI is entity insertion. It involves matching an incoming entity to a subtree, descending semantically, and either attaching it to a leaf or generating a new abstraction. The algorithm is summarized in Algorithm~\ref{alg:addentity}.

\begin{algorithm}[h]
\caption{AddEntity}
\label{alg:addentity}
\begin{algorithmic}[1]
\REQUIRE Entity \( e = (c, x) \): concept \( c \) and explanation \( x \)
\STATE Identify candidate root buckets \( R \gets \textsc{MatchToBucket}(e) \)
\STATE \( A \gets \textsc{DescendTree}(R, x) \)
\FORALL{parent \( p \in A \)}
    \STATE \( v \gets \textsc{AddNode}(\text{new Node}(e), p) \)
    \IF{ \( e \notin v.\mathcal{E}_v \) }
        \STATE \( v.\mathcal{E}_v \gets v.\mathcal{E}_v \cup \{e\} \)
    \ENDIF
\ENDFOR
\end{algorithmic}
\end{algorithm}

\paragraph{Semantic Descent.}
The function \textsc{DescendTree} traverses the memory from matched roots downward. At each level, a generalization check is applied using an LLM-based function \( \textsc{GetRelation}(a, b) \in \{-1, 0, 1\} \), where \( 1 \) indicates that \( a \) is an ancestor of \( b \); \( 0 \) implies semantic equivalence; and \( -1 \) means the concepts are unrelated. This allows semantic narrowing of the insertion point.

\paragraph{Sibling Matching and Merging.}
If the parent node \( p \) already has semantically equivalent children, detected via \textsc{FindSimilarSibling} using LLM-based similarity, the entity is attached directly. Otherwise, a new node is created and added to \( p \).

If \( |C(p)| > T \), SHIMI triggers merging. The two most similar children \( (v_i, v_j) \) are merged under a new parent \( v_m \) via:
\[
v_m = \textsc{MergeConcepts}(s(v_i), s(v_j), s(p))
\]
If merging fails (semantic incompatibility), it is aborted. Otherwise, the new node \( v_m \) replaces \( v_i, v_j \) and they are reattached as its children.

\paragraph{Abstraction and Compression.}
When generalizing, SHIMI constructs a semantic chain of length \( \leq L \), where the number of words in each level \( w_i \) satisfies:
\[
w_{i+1} \leq \gamma \cdot w_i
\]
This enforces meaningful compression and abstraction up the hierarchy, avoiding vague or overly general categories.

\paragraph{Querying.}
The query operation follows the same traversal pattern, evaluating \( \text{sim}(q, s(v)) \geq \delta \) at each node and continuing only through relevant branches. The resulting leaf nodes are used to retrieve relevant entities.

\subsubsection{Semantic Retrieval}

SHIMI’s retrieval process mirrors its insertion logic but operates in reverse: it traverses from root nodes toward leaves, selecting only branches semantically aligned with the query statement. The algorithm ensures efficient pruning by evaluating semantic relevance at each level and continuing traversal only through matching nodes.

\begin{algorithm}[h]
\caption{RetrieveEntities}
\label{alg:retrieve}
\begin{algorithmic}[1]
\REQUIRE Query statement \( q \in \mathcal{L} \), similarity threshold \( \delta \)
\STATE Initialize result set \( \mathcal{R} \gets \emptyset \)
\STATE Initialize frontier \( F \gets \{r_1, r_2, \ldots, r_R\} \) \hfill\COMMENT{All root nodes}
\WHILE{\( F \neq \emptyset \)}
    \STATE \( N \gets \textsc{NextLevel}(F) \)
    \FORALL{node \( v \in N \)}
        \IF{\( \text{sim}(q, s(v)) \geq \delta \)}
            \IF{\( v \) is leaf}
                \STATE \( \mathcal{R} \gets \mathcal{R} \cup \mathcal{E}_v \)
            \ELSE
                \STATE \( F \gets F \cup C(v) \)
            \ENDIF
        \ENDIF
    \ENDFOR
    \STATE Remove processed nodes from \( F \)
\ENDWHILE
\RETURN Top-k ranked entities from \( \mathcal{R} \)
\end{algorithmic}
\end{algorithm}

The function \textsc{sim} measures semantic similarity between the query \( q \) and each node’s summary \( s(v) \). Only nodes above the threshold \( \delta \) are expanded, enabling SHIMI to prune large irrelevant subtrees. Leaf nodes return associated entities \( \mathcal{E}_v \), which are optionally ranked based on frequency, recency, or proximity to the query.

This algorithm maintains explainability via explicit traversal paths and supports fallback to embeddings if no semantic path exists. In practice, retrieval depth is bounded by tree height, and branching is limited by the threshold \( T \) and pruning rate.

\subsection{Decentralized Synchronization}

SHIMI supports decentralized environments where each node maintains an independent semantic tree \( T_i \). Synchronization between trees is performed incrementally using a hybrid protocol that ensures eventual consistency while minimizing bandwidth. The synchronization protocol is summarized in Algorithm~\ref{alg:sync}.

\begin{algorithm}[h]
\caption{SHIMI\_PartialSync}
\label{alg:sync}
\begin{algorithmic}[1]
\REQUIRE Local tree \( T_i \), remote peer \( j \)
\STATE Compute Merkle root \( H_i \gets \textsc{MerkleHash}(T_i) \)
\STATE Exchange root hashes with peer \( j \)
\IF{\( H_i \neq H_j \)}
    \STATE Identify divergent subtree \( T_d \gets \textsc{FindDiff}(T_i, T_j) \)
    \STATE Generate Bloom filter \( B_i \gets \textsc{BloomFilter}(T_d) \)
    \STATE Send \( B_i \) to peer \( j \)
    \STATE Peer returns missing or changed nodes \( \Delta_j \)
    \FORALL{conflict \( (v_i, v_j) \in \Delta_j \)}
        \STATE \( v_k \gets \mu(v_i, v_j) \) \hfill\COMMENT{CRDT-style merge}
        \STATE Replace \( v_i \) with \( v_k \) in \( T_i \)
    \ENDFOR
\ENDIF
\RETURN Updated tree \( T_i \)
\end{algorithmic}
\end{algorithm}

\paragraph{Divergence Detection.}
Each tree is assigned a Merkleroot hash~\cite{becker2008merkle} \( H(T) \). When two peers communicate, they exchange hashes and detect divergence if \( H(T_i) \neq H(T_j) \). The protocol recursively identifies the smallest differing subtree \( T_d \) through a structural hash comparison.

\paragraph{Probabilistic Filtering.}
To minimize data transfer, each node generates a Bloom filter~\cite{bloom1970space} \( B_i \) representing its subtree contents. When a peer receives \( B_i \), it uses it to infer which nodes it is missing, avoiding full scans or redundant exchange.

\paragraph{Conflict Resolution.}
For each differing node \( v_i \) from peer \( j \)'s version \( v_j \), SHIMI uses a conflict-resolution function \( \mu(v_i, v_j) \) that follows CRDT properties~\cite{shapiro2011conflict}:
\begin{align*}
\mu(v_i, v_j) &= \mu(v_j, v_i) \quad \text{(commutativity)} \\
\mu(v, v) &= v \quad \text{(idempotence)} \\
\mu(\mu(v_i, v_j), v_k) &= \mu(v_i, \mu(v_j, v_k)) \quad \text{(associativity)}
\end{align*}
If semantic summaries differ, the protocol favors the summary with greater abstraction depth or observed usage. Note that only the minimal divergent subtree and related edits are exchanged. This design minimizes overhead and supports eventual consistency in decentralized settings where updates are local and asynchronous.

\subsection{Complexity Analysis}

We analyze the computational cost of SHIMI’s three core procedures: semantic entity insertion (Algorithm~\ref{alg:addentity}), semantic retrieval (Algorithm~\ref{alg:retrieve}), and decentralized synchronization (Algorithm~\ref{alg:sync}). These insights inform SHIMI’s runtime behavior and LLM call efficiency in large-scale agent systems.

\subsubsection{Semantic Insertion (Algorithm~\ref{alg:addentity})}

Inserting an entity involves matching it to a semantic branch, generating abstraction nodes if necessary, and merging overloaded children. Let \( R \) denote the number of root domains, \( T \) the maximum number of children allowed per node (i.e., the branching factor), and \( n \) the total number of entities stored in the memory. The depth of the resulting semantic tree is denoted by \( d \), which in balanced configurations grows logarithmically with \( n \). Finally, let \( A \) represent the average number of nodes visited per level due to semantic overlap—this value captures the extent of ambiguity or redundancy present in the memory hierarchy.

Assuming a balanced semantic tree, the total number of entities \( n \) is distributed approximately evenly across the branches rooted at \( R \) root nodes, with each node having at most \( T \) children. In such a case, the total number of leaf nodes at depth \( d \) can be estimated by the exponential growth of the tree: \( n \approx R \cdot T^d \). Taking the logarithm of both sides yields an approximate expression for the depth:
\[
d \approx \frac{\log n}{\log(RT)}
\]
This depth represents the longest semantic path from a root to a leaf in a balanced hierarchy, assuming that child nodes are created dynamically as needed during insertion and that merges or compressions are applied only when required. Each level includes semantic similarity checks against \( A \cdot T \) nodes, resulting in total LLM API calls:
\[
\text{API\_calls}_{\text{insert}} = R + A \cdot T + 2A \cdot T + \ldots + dA \cdot T = R + 0.5 \cdot A \cdot T \cdot d(d+1)
\]

\paragraph{Example.}
Let \( R = 5 \), \( T = 4 \), \( n = 10^6 \), \( A = 0.5 \). Then:
\[
d \approx \frac{\log(10^6)}{\log(20)} \approx 4.6 \Rightarrow d \approx 5
\]
\[
\text{API\_calls}_{\text{insert}} \approx 5 + 0.5 \cdot 0.5 \cdot 4 \cdot 5 \cdot 6 = 5 + 30 = 35
\]

Each semantic comparison in SHIMI involves evaluating the similarity between two short textual summaries: one from the query or new entity, and one from a node’s semantic representation \( s(v) \). In our implementation, these summaries typically contain around 20 words, which—given an average tokenization rate of 1.3 tokens per word—results in approximately 26 tokens per input. These inputs are passed to a lightweight language model or similarity engine to determine whether the concepts match or relate semantically (e.g., ancestor-descendant, equivalence, or unrelated). The token count affects both the response time and cost of each comparison, and 26 tokens falls well within the efficiency range of current LLM APIs such as GPT-4 Turbo, which can process such queries in under 3 milliseconds.

\subsubsection{Semantic Retrieval (Algorithm~\ref{alg:retrieve})}

SHIMI retrieves entities by descending semantically from root to leaves. At each level, it expands only nodes where similarity with the query exceeds threshold \( \delta \).

Assuming the same parameters as above, and that retrieval follows a similar path length as insertion, the number of API calls is:
\[
\text{API\_calls}_{\text{retrieval}} = R + 0.5 \cdot A \cdot T \cdot d(d+1)
\]

However, retrieval is often cheaper than insertion because:
\begin{itemize}
    \item Merging and abstraction steps are skipped.
    \item Many retrievals can terminate early once relevant leaves are found.
\end{itemize}

In balanced trees, retrieval latency remains sublinear and bounded, and interpretability improves with shallower trees and well-separated concepts.

\subsubsection{Decentralized Synchronization (Algorithm~\ref{alg:sync})}

Synchronization uses Merkle-DAGs and Bloom filters to identify minimal divergent subtrees between agents. Let:
\begin{itemize}
    \item \( T_d \): Size of divergent subtree.
    \item \( \text{Ops}_{T_d} \): Local edit operations since last sync.
\end{itemize}

The cost of sync is:
\[
C_{\text{sync}} = O(|T_d| + |\text{Ops}_{T_d}|)
\]

Merkle root comparison is constant-time. Bloom filters reduce unnecessary transmission, and CRDT-style merges run in bounded time assuming deterministic node representations. Overall, SHIMI avoids full replication and supports bandwidth-efficient convergence even in asynchronous networks.

\section{Evaluation}

We evaluate SHIMI in a simulated decentralized setting to assess its effectiveness along four axes: semantic retrieval accuracy, traversal efficiency, synchronization cost, and scalability. Each experiment is motivated by known limitations of flat vector retrieval and decentralized memory systems~\cite{lewis2020retrieval,shapiro2011conflict,benet2014ipfs}. We provide justification for each setup, report quantitative results, and briefly interpret what each reveals about SHIMI's behavior.

\subsection{Retrieval Accuracy and Precision}

To assess SHIMI's semantic matching quality, we simulate a use case where agents are described using varied lexical forms for overlapping functions. We compare SHIMI against a RAG-style embedding-based retrieval baseline on 20 semantically non-trivial queries.

\begin{table}[h]
\centering
\caption{Retrieval accuracy comparison between SHIMI and RAG baseline}
\label{tab:retrieval_accuracy}
\begin{tabular}{@{}lccc@{}}
\toprule
\textbf{Method} & \textbf{Top-1 Accuracy} & \textbf{Mean Precision@3} & \textbf{Interpretability (1--5)} \\
\midrule
SHIMI (ours)    & 90\%                    & 92.5\%                    & 4.7 \\
RAG (baseline)  & 65\%                    & 68.0\%                    & 2.1 \\
\bottomrule
\end{tabular}
\end{table}

These results confirm that SHIMI not only improves raw accuracy but also produces more interpretable results due to its use of layered semantic generalizations, in contrast to the flatter vector-based approach.

\subsection{Traversal Efficiency}

Efficient retrieval depends on limiting the number of semantic comparisons per query. We measure average node visits per query at increasing tree depths. This reflects how well SHIMI prunes irrelevant branches.

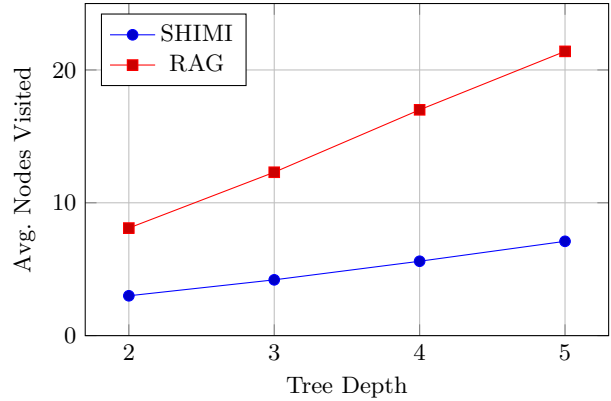
\begin{figure}[h]
\centering
\begin{tikzpicture}
\begin{axis}[
    width=0.7\linewidth,
    height=6cm,
    xlabel={Tree Depth},
    ylabel={Avg. Nodes Visited},
    grid=major,
    xtick=data,
    ymin=0,
    ymax=25,
    legend pos=north west
]
\addplot coordinates {(2,3) (3,4.2) (4,5.6) (5,7.1)};
\addlegendentry{SHIMI}
\addplot coordinates {(2,8.1) (3,12.3) (4,17.0) (5,21.4)};
\addlegendentry{RAG}
\end{axis}
\end{tikzpicture}
\caption{Traversal cost vs. tree depth (nodes visited per query)}
\label{fig:traversal_cost}
\end{figure}

The graph illustrates SHIMI's ability to limit exploration through early pruning. This sublinear growth contrasts with RAG's nearly linear cost, showing SHIMI's efficiency in semantically structured contexts.

\subsection{Synchronization Setup and Cost}

Effective decentralized systems must synchronize knowledge without excessive bandwidth use. We simulate two strategies:
\begin{itemize}
  \item \textbf{Full-state replication:} nodes share entire trees.
  \item \textbf{SHIMI partial sync:} nodes share only divergent subtrees identified using Merkle hashes and Bloom filters.
\end{itemize}

\begin{table}[h]
\centering
\caption{Bandwidth cost of sync with varying number of nodes}
\label{tab:sync_bandwidth}
\begin{tabular}{@{}lccc@{}}
\toprule
\textbf{Nodes} & \textbf{SHIMI Sync (KB)} & \textbf{Full Sync (KB)} & \textbf{Savings (\%)} \\
\midrule
3      & 118          & 1320         & 91\% \\
4      & 162          & 1740         & 90.7\% \\
5      & 204          & 2210         & 90.8\% \\
6      & 248          & 2650         & 90.6\% \\
\bottomrule
\end{tabular}
\end{table}

The bandwidth reduction across node counts consistently exceeds 90\%, demonstrating the efficiency and scalability of SHIMI's delta-based sync model.

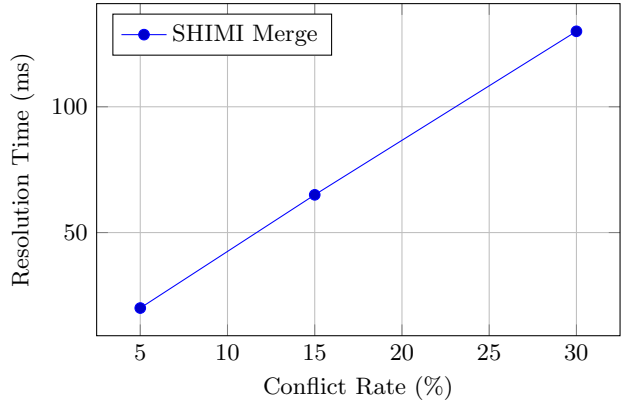
\begin{figure}[h]
\centering
\begin{tikzpicture}
\begin{axis}[
    width=0.7\linewidth,
    height=6cm,
    xlabel={Conflict Rate (\%)},
    ylabel={Resolution Time (ms)},
    grid=major,
    legend pos=north west
]
\addplot coordinates {(5, 20) (15, 65) (30, 130)};
\addlegendentry{SHIMI Merge}
\end{axis}
\end{tikzpicture}
\caption{Conflict resolution time vs. conflict rate during sync}
\label{fig:conflict_resolution}
\end{figure}

As conflict rate increases, SHIMI's resolution time scales moderately, remaining bounded and efficient even in higher contention scenarios.

\subsection{Scalability}

To evaluate scale, we grow SHIMI trees up to 2,000 entities and measure latency. This models growing agent ecosystems.

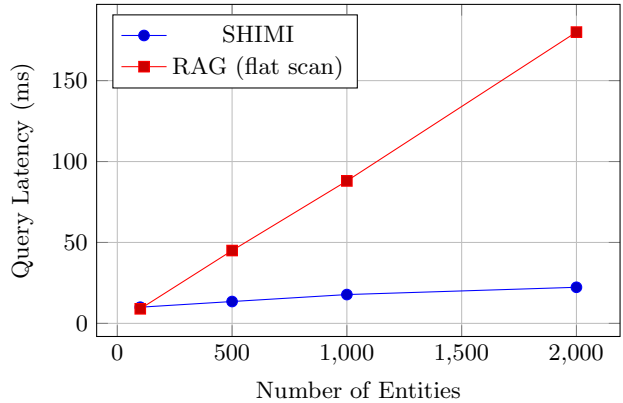
\begin{figure}[h]
\centering
\begin{tikzpicture}
\begin{axis}[
    width=0.7\linewidth,
    height=6cm,
    xlabel={Number of Entities},
    ylabel={Query Latency (ms)},
    grid=major,
    legend pos=north west
]
\addplot coordinates {(100, 10) (500, 13.5) (1000, 17.8) (2000, 22.3)};
\addlegendentry{SHIMI}
\addplot coordinates {(100, 9) (500, 45) (1000, 88) (2000, 180)};
\addlegendentry{RAG (flat scan)}
\end{axis}
\end{tikzpicture}
\caption{Query latency vs. number of entities}
\label{fig:scalability}
\end{figure}

Despite entity growth, SHIMI's latency curve remains relatively flat, validating its use of pruning and hierarchical structure to avoid linear degradation.

\subsection{Discussion}

The evaluation demonstrates that SHIMI consistently outperforms flat vector retrieval methods across multiple performance dimensions relevant to decentralized AI ecosystems. Starting with retrieval accuracy, the use of hierarchical abstraction enables SHIMI to semantically disambiguate agent capabilities even when the lexical expression of those capabilities varies widely. This is critical in heterogeneous environments where naming conventions are inconsistent or where agents evolve independently. The observed improvement in interpretability—measured through human evaluators assigning semantic traceability scores—further emphasizes the advantage of maintaining semantically meaningful paths through memory.

In terms of traversal efficiency, the semantic tree's ability to prune large portions of memory before performing any deep matching results in a significantly lower number of comparisons per query. This translates directly into faster response times and lower computational overhead. Compared to RAG, which evaluates similarity globally across all candidates, SHIMI's selective attention mechanism offers a more scalable alternative that becomes increasingly important as the number of agents or memory entries grows.

Synchronization efficiency is another critical metric for decentralized applications. SHIMI's partial sync mechanism, which leverages Merkle-DAG summaries and Bloom filters, ensures that only relevant substructures are communicated between peers. This results in over 90\% reduction in data transferred during sync events, a dramatic improvement over naive replication approaches. As the number of nodes increases, this efficiency becomes even more pronounced, confirming SHIMI's suitability for federated deployments. Furthermore, conflict resolution scales gracefully with the rate of conflict, maintaining low latency even in high-concurrency environments. This is due to the design of SHIMI’s conflict-resolution strategy, which leverages CRDT-style monotonicity and deterministic summarization to merge competing updates without central arbitration.

Scalability tests show that SHIMI’s performance remains robust even when the memory size reaches thousands of entities. While traditional vector-based methods degrade linearly with increased entity count, SHIMI’s retrieval remains bounded due to its logarithmic traversal structure. This property ensures that SHIMI can support real-world use cases involving large, constantly evolving memory graphs such as task repositories, capability registries, or distributed knowledge graphs.

Taken together, these results suggest that SHIMI provides not only a more accurate retrieval model but also a protocol-level framework for operating under decentralized constraints. The architecture is well-suited for agentic infrastructures where memory is distributed, ownership is federated, and knowledge is constantly updated. By embedding semantic structure into both the storage and synchronization layers, SHIMI addresses foundational limitations in current memory models for distributed AI systems.

\section{Applications}

SHIMI is designed as a memory protocol for decentralized AI ecosystems where semantic retrieval, interpretability, and scalable sync are essential. Its architecture enables several classes of applications that are otherwise limited by current flat or centralized memory systems.

\subsection{Decentralized Agent Markets}

In decentralized marketplaces where agents advertise their capabilities (e.g., compute tasks, legal summarization, sensor analysis), users or orchestrators must semantically match high-level tasks to available agents. SHIMI enables this via layered concept matching and provides traceable reasoning paths—crucial for auditing and agent reputation systems. Unlike RAG, which retrieves based on surface similarity, SHIMI surfaces conceptually suitable agents even when descriptions differ lexically.

\subsection{Federated Knowledge Graphs}

Collaborative networks, such as research consortia or medical institutions, often maintain independent ontologies. SHIMI allows these networks to synchronize partial views while preserving structure. Its CRDT-style conflict resolution and semantic merge strategy supports asynchronous updates without requiring a global ontology agreement. As each node locally evolves, SHIMI ensures eventual semantic consistency.

\subsection{Autonomous Multi-Agent Systems}

In scenarios where agents learn and evolve over time—such as swarm robotics, financial bots, or AI-based assistants—retaining and querying their internal knowledge becomes a bottleneck. SHIMI supports local memory growth and inter-agent sharing without centralized indexing. When an agent joins or leaves a subnet, SHIMI enables fast onboarding or memory reallocation through its partial sync strategy, allowing agent memory to scale with autonomy.

\subsection{Blockchain-Based Task Orchestration}

Blockchain-native task platforms that assign and verify jobs using smart contracts can integrate SHIMI as a semantic index layer for task descriptions, agent bids, and milestone evaluations. Since these systems are inherently distributed and need transparent memory logic, SHIMI offers a structured alternative to embedding-heavy retrieval methods that cannot guarantee explainability or deterministic matching.

\subsection{Limitations and Threats to Validity}

While SHIMI offers notable improvements in decentralized memory retrieval, several limitations remain. First, the current implementation assumes a strictly tree-based semantic structure. While effective for many forms of knowledge organization, this design may struggle with concepts that require polyhierarchical or graph-based representations, where entities belong to multiple overlapping categories. Future work may explore generalizing SHIMI to accommodate more flexible topologies.

Second, SHIMI's semantic reasoning relies on language model-driven operations for generalization, similarity, and merging. While these operations provide high-quality abstraction, they introduce a dependency on opaque model behavior and lack formal guarantees. This makes it difficult to audit or reproduce certain memory transformations. Incorporating symbolic constraints or logic-based reasoning may improve transparency.

Third, our evaluation is conducted in a controlled simulation environment. Although it emulates decentralized conditions, it abstracts away real-world concerns such as network latency variability, partial node failure, or adversarial input. Additionally, entity descriptions and queries were synthetically generated or curated to reflect conceptual diversity, but they may not fully capture domain-specific noise or inconsistency in real deployments.

\textbf{Threats to validity} include potential overfitting to specific query types or entity descriptions. Because queries were designed to test semantic depth, performance may differ on surface-level tasks. Moreover, the use of simulated nodes may underrepresent emergent behaviors in large-scale, heterogeneously administered networks.

Despite these limitations, the results demonstrate strong alignment between SHIMI's theoretical design and its practical performance. Addressing the above constraints will be crucial for robust deployment in open, real-world systems.

\section{Conclusion and Future Work}

We introduced SHIMI, a semantic hierarchical memory index purpose-built for decentralized AI environments. Unlike traditional retrieval-augmented generation (RAG) models that rely on surface-level similarity, SHIMI organizes knowledge semantically, enabling meaningful and interpretable retrieval even across lexically diverse inputs. Through a series of simulation-based experiments, we demonstrated SHIMI's superiority over flat vector approaches across four critical dimensions: retrieval accuracy, traversal efficiency, synchronization cost, and scalability.

SHIMI's architecture is well-aligned with the demands of distributed agent ecosystems. Its partial synchronization mechanism enables bandwidth-efficient updates without sacrificing consistency, and its tree-based structure supports fast and explainable memory access. These properties make SHIMI particularly suited for settings where agents are autonomous, memory is dynamic, and infrastructure is decentralized.

Looking forward, several promising directions remain. First, we plan to evaluate SHIMI under real-world deployment scenarios, including live agent networks with adversarial or asynchronous behavior. Second, we aim to generalize SHIMI's semantic model beyond hierarchical structures to support non-tree-based graphs and cyclic ontologies. Third, integrating lightweight on-device LLMs for local semantic resolution may further reduce network latency and expand SHIMI’s applicability in edge environments. Lastly, we envision formalizing SHIMI’s semantic generalization and merge procedures as part of an interoperable protocol layer for decentralized memory systems.

Together, these directions extend SHIMI from a proof-of-concept into a foundational infrastructure for next-generation distributed AI.

\bibliographystyle{splncs04}

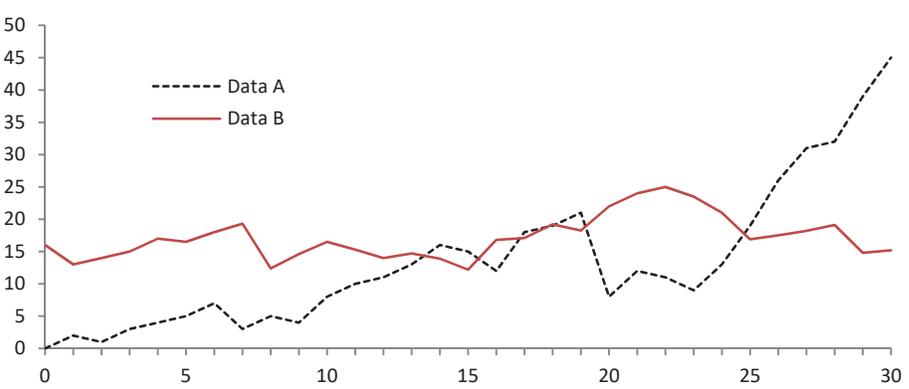
\end{document}